\documentclass{article}
\usepackage{ijcai26}

\usepackage{times}
\usepackage{soul}
\usepackage{url}
\usepackage[hidelinks]{hyperref}
\usepackage[utf8]{inputenc}
\usepackage[small]{caption}
\usepackage{graphicx}
\usepackage{amsmath}
\usepackage{amsthm}
\usepackage{booktabs}
\usepackage{algorithm}
\usepackage{algorithmic}
\usepackage[switch]{lineno}

\makeatletter
\def\blfootnote{\gdef\@thefnmark{}\@footnotetext}
\makeatother

\title{VFEAgent: A Multimodal Agent Framework \\for End-to-End Automated Finite Element Analysis}

\author{
Jiachen Zhang$^{1,2,*}$
\and
Junyi Lao$^{1,*}$\and
Chenghao Liu$^{1,*}$\and
Siyuan Liu$^1$\and\\
Shixin Wu$^1$\and
Linsen Zhang$^1$\and
Boyu Wang$^1$\And
Songfang Huang$^{1,\dagger}$\\
\affiliations
$^1$Peking University\\
$^2$China Agricultural University\\
}
\begin{document}

\maketitle
\blfootnote{\textsuperscript{*}These authors contributed equally to this work.}
\blfootnote{\textsuperscript{$\dagger$}Corresponding author \textit{E-mail address:} hsf@pku.edu.cn}

\begin{abstract}
    Finite Element Analysis (FEA) serves as the cornerstone of modern engineering design. However, its workflow is inherently complex and relies heavily on domain expertise. Although recent efforts have integrated Large Language Models (LLMs) into FEA, existing approaches face limitations in handling multimodal inputs and executing complex tasks. To address these limitations, we propose VFEAgent, an end-to-end multi-agent system designed to automate FEA modeling and simulation directly from input images and problem descriptions.
    
    Our methodology integrates two core components: (1) a multimodal vision-language multi-agent pipeline that employs ReAct-driven reasoning to extract structured FEA specifications from heterogeneous inputs and (2) a verification-first code synthesis framework, incorporating robust self-debugging and fallback mechanisms to ensure executability and physical validity. We systematically evaluated the system across various engineering mechanics scenarios. The results demonstrate that VFEAgent achieves a high success rate in generating complete and physically valid simulations, outperforming LLM-based baseline methods in reliability and correctness. These findings validate the feasibility of automating the complete FEA workflow, highlighting the framework's potential to liberate engineers from tedious manual analysis.
\end{abstract}

\section{Introduction}
Structural engineering analysis is the cornerstone of ensuring the safety and reliability of large-scale infrastructure, aerospace vehicles, and complex mechanical assemblies, directly determining the scientific validity of engineering decisions. As the primary methodology of computational mechanics, Finite Element Analysis (FEA) has fundamentally transformed the paradigm of structural analysis by discretizing continuous media into numerical elements \cite{belytschko2022eighty}. However, despite the formidable power of contemporary computational engines, the current FEA workflow remains characterized by high empirical dependency and intensive human-computer interaction \cite{sahani2024aienhanced}. This necessitates that expert engineers manually bridge the chasm between engineering drawings, geometric modeling, and physical environment configuration \cite{chen2025critical}. Such a process is not only labor-intensive and time-consuming but also highly susceptible to human error during tedious parameter configuration and topological processing \cite{shah2024application}. In the context of modern engineering demanding high-frequency iterations, this over-reliance on manual operation constrains the deep integration of digital design and intelligent manufacturing.

The rapid ascent of Large Language Models (LLMs) and Vision-Language Models (VLMs) offers a transformative opportunity to automate the FEA pipeline \cite{baker2025large}. By leveraging the scripting interfaces of mainstream commercial software (e.g., the Abaqus Scripting Interface/API), Multi-Agent Systems (MAS) can theoretically translate high-level engineering intents into precise instruction sequences \cite{ni2024mechagents,hong2024metagpt}. Nevertheless, existing AI-integrated FEA frameworks face three critical limitations that hinder their practical utilit \cite{sahani2024aienhanced}. Most fundamentally, \textbf{existing works predominantly rely on hard-coded pipelines or preset knowledge bases.} Instead of performing genuine de novo geometric construction, these systems execute pseudo-modeling" by populating parameters into prototypes retrieved from pre-defined libraries \cite{qi2025feagpt}. Such template-matching fails to satisfy the vast majority of real-world design requirements, which demand precise, customized parametric representations of non-standard structures under non-fixed operating conditions \cite{gopfert2023opportunities}. Compounding this issue, \textbf{current agents are often vision-blind"}. By utilizing pre-processed structured text or meshes as input, they circumvent the most challenging phase of FEA—the interpretation of raw blueprints—thereby losing high-fidelity semantic information \cite{khan2024finetuning}. Furthermore, \textbf{feedback mechanisms remain immature.} Existing debugging strategies are largely confined to shallow syntax errors, lacking the ability to identify implicit physical logic discrepancies or form a closed-loop self-healing capability that integrates long-term experience with short-term reflection \cite{chen2024parameters,shinn2024reflexion}.

\begin{figure*}[t]
    \centering
    \includegraphics[width=1\linewidth]{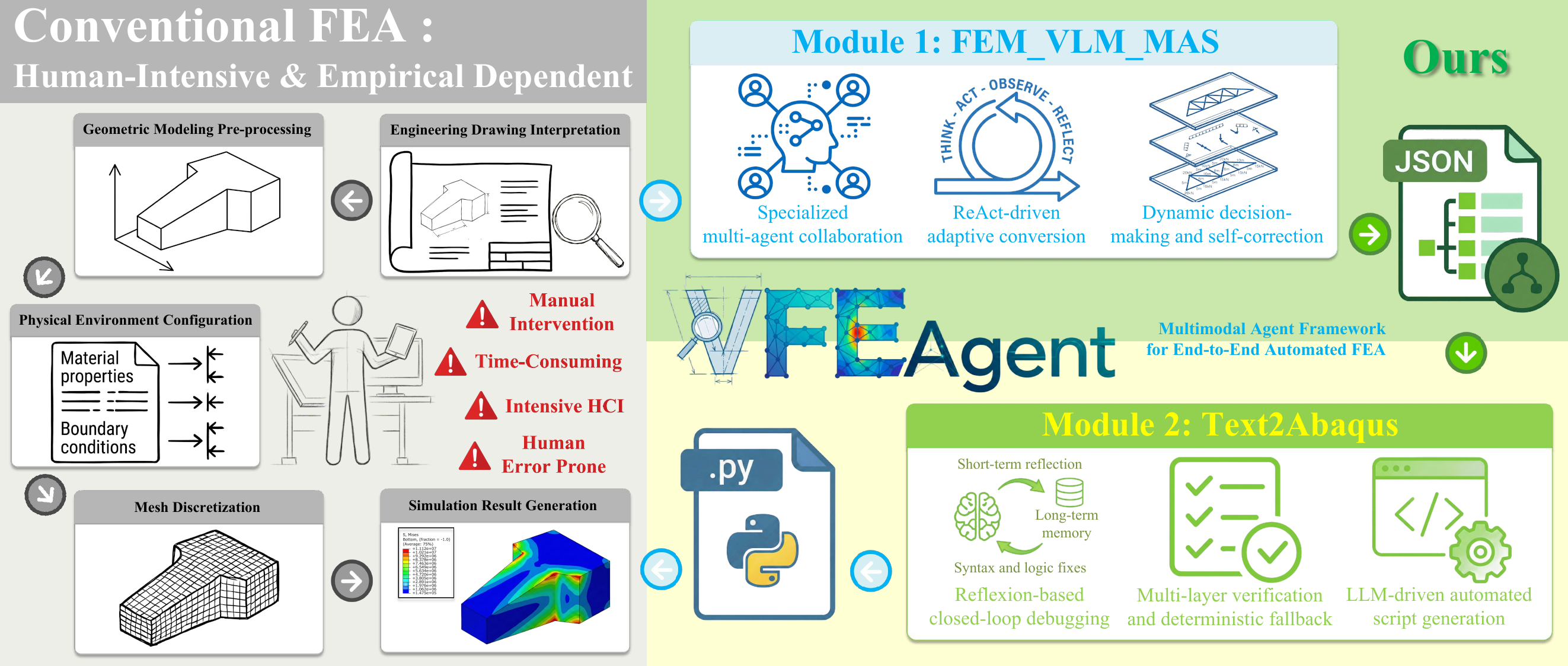}
    \caption{\textbf{VFEAgent automates the transformation of engineering drawings into validated finite element simulation results.}}
    \label{fig1}
\end{figure*}

To address these gaps and explore the profound potential of AI in autonomous engineering, we propose VFEAgent, an end-to-end framework validated on the commercial software Abaqus \cite{hou2025autofea}. VFEAgent is the first autonomous FEA framework capable of transforming raw engineering drawings into executable simulation scripts and computational results \cite{qi2025feagpt}. The system is driven by two synergistic components: FEM\_VLM\_MAS, which employs a ReAct-driven hierarchical perception strategy to parse blueprints, and Text2Abaqus, an LLM-driven engine that generates and executes simulation scripts \cite{ni2024mechagents,yao2023react,wei2022chain}. Unlike prior pseudo-modeling attempts, VFEAgent decodes visual signals into structured engineering semantics and synthesizes models from the ground up \cite{chen2025from}. To bridge the gap in specialized evaluation, we further introduce a graded expert-curated, vision-augmented benchmark, containing 15 complex cases—ranging from asymmetrical steel frames and beam topology optimization to pressure vessels and metamaterials—to evaluate core physical quantities such as Mises stress, modal frequencies, acceleration, and strain energy \cite{mohammadzadeh2025fembench}.
The main contributions of this paper are as follows:
\begin{itemize}
\item \textbf{Realization of the first genuinely end-to-end FEA multi-agent system.} Diverging from retrieval-based pseudo-modeling, VFEAgent achieves full-process autonomy—from raw image input to geometric construction, physical solving, and post-processing—demonstrating strong universality for non-standard, complex industrial scenarios.
\item \textbf{A Hierarchical Perception Mechanism.} By simulating expert habits, FEM\_VLM\_MAS system utilizes a multi-level extraction strategy to decode engineering drawings. It progressively decodes engineering drawings from numerical values and nodes to load distributions and connectivity, enabling precise transformation from visual signals to structured engineering semantics.
\item \textbf{Self-Healing Script Generation Engine.} By introducing a debugger that fuses short-term reflection with long-term memory, Text2Abaqus system enables closed-loop rectification of both syntax errors and physical logic biases, significantly enhancing the reliability of autonomous simulation.
\item \textbf{Release of a Graded Solid Mechanics Evaluation Framework.} We provide an expert-validated vision-augmented benchmark featuring complex physical scenarios and multi-dimensional metrics. This framework serves as a quantitative baseline for evaluating the engineering design thinking and constraint-processing capabilities of large models.
\end{itemize}

\section{Related Work}
\subsection{LLM-Integrated FEA}
Early research on integrating Large Language Models (LLMs) into Finite Element Analysis (FEA) primarily focused on automating individual stages of the simulation lifecycle to reduce manual labor. In geometry and meshing, NekMesh \cite{green2024nekmesh} introduced a CAD-independent framework for high-order mesh generation using variational untangling and octree-based adaptive sampling. For expert-level decision support, Gembarski \cite{gembarski2020agent} proposed a multi-agent system (MAS) that simulates expert negotiations to identify model features and manufacturing constraints. Recently, frameworks like FRAME \cite{guru2025frame} and the work by Tian et al. \cite{tian2024optimizing} explored ``self-optimization'' workflows that identify physical issues, such as stress concentration, and automatically trigger geometric modifications. However, these partial assistants typically require structured inputs and human intervention to bridge the chasm between disparate simulation phases.

\begin{figure*}[t]
    \centering
    \includegraphics[width=0.95\linewidth]{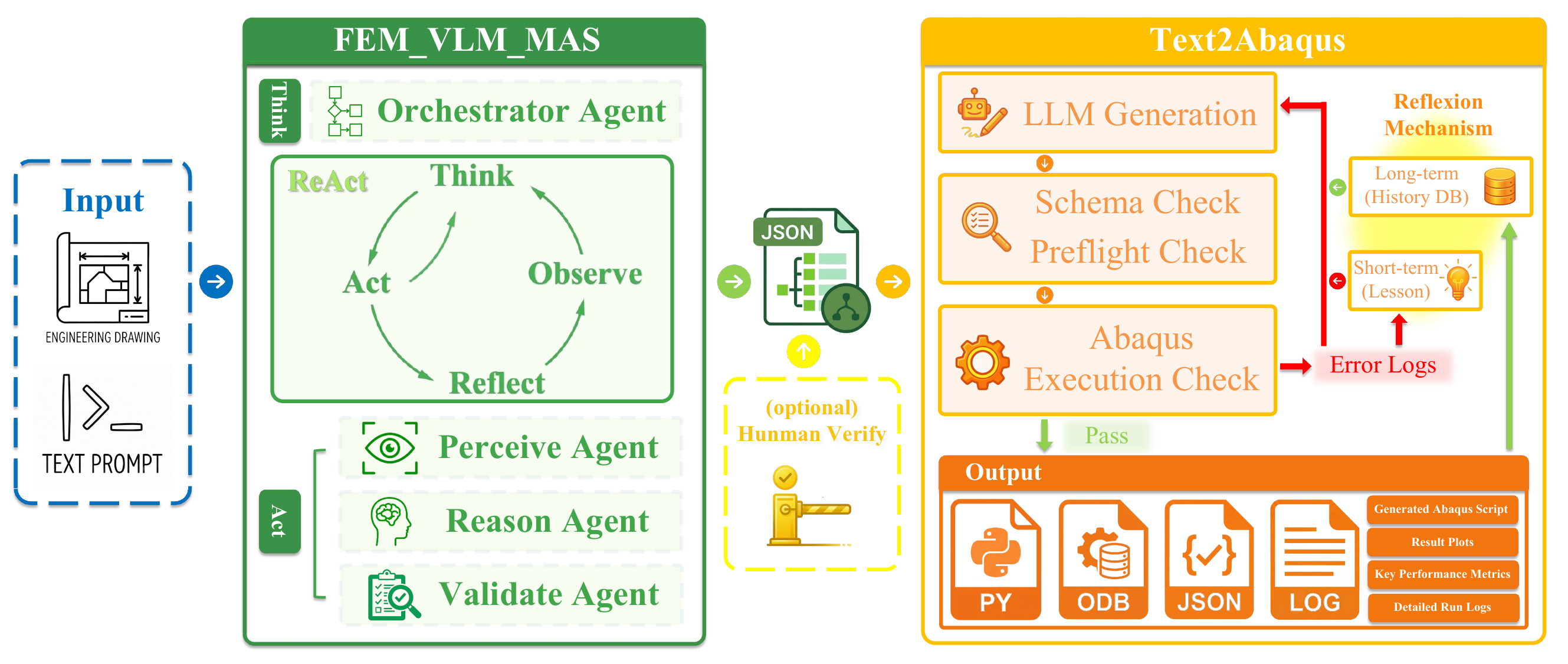}
    \caption{\textbf{The Neuro-Symbolic Architecture of VFEAgent.} The framework bridges the semantic gap between visual diagrams and physical simulation via two coupled stages: \textbf{(A) Perception}, employing a multi-agent ReAct system to extract a solver-agnostic Intermediate Representation (IR); and \textbf{(B) Synthesis}, featuring a verification-driven loop that integrates AST-based preflight checks, reflexive debugging, and a deterministic handover protocol to guarantee executability.}
    \label{fig:overview}
\end{figure*}

The field has recently evolved toward end-to-end autonomous agents that orchestrate the full ``Geometry-Mesh-Simulation-Analysis'' (GMSA) pipeline. MechAgents \cite{ni2024mechagents} utilized role-playing agents to solve elasticity problems using FEniCS, while FeaGPT \cite{qi2025feagpt} implemented the first conversational GMSA pipeline driven by natural language. Furthermore, Geng et al. \cite{geng2025lightweight} proposed a lightweight MAS framework to automate 2D frame modeling by decoupling structural analysis into specialized tasks. In fluid mechanics, ChatCFD \cite{fan2025chatcfd} leveraged structural reasoning to correct complex coupled errors in compressible flows. Despite these advancements, existing frameworks heuristically bypass core challenges through ``pseudo-modeling'' strategies. Systems such as MooseAgent \cite{zhang2025mooseagent} and AutoFEA \cite{hou2025autofea} rely heavily on Retrieval-Augmented Generation (RAG) to fill parameters into predefined templates rather than performing genuine \textit{de novo} construction. Furthermore, these models remain largely ``vision-blind,'' bypassing the critical interpretation of raw engineering blueprints and struggling with deep physical logic, such as non-linear plasticity.

\subsection{Evaluation Benchmarks}
Evaluation benchmarks have similarly lagged behind the demands of autonomous engineering. Current benchmarks primarily focus on either commercial API operation, such as FEABench \cite{anonymous2025feabench}, or underlying mathematical implementation, such as FEM-BENCH \cite{mohammadzadeh2025fembench}. While EngDesign \cite{guo2025engdesign} introduced design-thinking metrics, its structural tasks are confined to idealized 2D geometries like simple trusses or rectangular beams. Overall, existing benchmarks lack coverage of non-standard industrial topologies and fail to assess the visual reasoning required to parse raw drawings. Moreover, they often focus on code syntax rather than high-fidelity physical field distributions, such as modal frequencies or strain energy. Consequently, a gap remains between passing synthetic code tests and solving real-world, vision-augmented engineering problems.

% ==================================================================
% SECTION 3: METHODOLOGY
% ==================================================================
\section{Methodology}

\subsection{Problem Formulation}
We define the automated FEA modeling task as a mapping from a multimodal input tuple $\mathcal{X} = (I, T_{ctx})$ to a valid physical response field $\mathcal{R}$. Here, $I$ denotes the structural diagram and $T_{ctx}$ contains textual constraints (e.g., material properties, load magnitudes). To ensure physical validity and software agnosticism, we factorize this end-to-end process into an intermediate grounded state $Y$:
\begin{equation}
\mathcal{X} \xrightarrow{\phi_{\text{perc}}} Y \xrightarrow{\phi_{\text{syn}}} S \xrightarrow{\text{Solver}} \mathcal{R}
\end{equation}
where $Y = (\mathcal{G}, \mathcal{M}, \mathcal{BC}, \mathcal{L})$ represents a \textit{solver-agnostic intermediate representation} (IR) enforced by a strict schema. Specifically, $\mathcal{G} = \{V, E\}$ denotes the geometry topology (nodes and connectivity), $\mathcal{M}$ specifies material sections, while $\mathcal{BC}$ and $\mathcal{L}$ define boundary conditions and load vectors mapped to $\mathcal{G}$. $S$ is the executable simulation script (e.g., Abaqus Python) generated from $Y$.

\subsection{System Architecture}
The \textbf{VFEAgent} framework implements the mappings $\phi_{\text{perc}}$ and $\phi_{\text{syn}}$ via two coupled modules. \textbf{Stage-A (FEM\_VLM\_MAS)} employs a multi-agent ReAct system to instantiate $Y$ from visual inputs. \textbf{Stage-B (Text2Abaqus)} implements a verification-driven synthesis loop with a neuro-symbolic handover protocol to guarantee the executability of $S$.

\subsection{Stage-A: ReAct-Driven Semantic Perception}
\paragraph{Multi-Agent Orchestration.} We decompose the complex visual reasoning task into four specialized roles: (1) \textit{Perception Agent} for OCR and geometric primitive detection; (2) \textit{Reasoning Agent} for inferring topological connectivity and component types; (3) \textit{Validation Agent} for auditing schema consistency; and (4) \textit{Orchestrator} for global planning. The Orchestrator maintains a dynamic belief state $B_t$ and executes a decision loop: $a_t = \pi(B_t, \text{Prompt}_{ctx})$, where actions $a_t$ include invoking sub-agents or updating the schema.

\subsubsection{Visual Perception Mechanism}
\paragraph{End-to-End Direct Reasoning.} Unlike pipeline approaches that rely on intermediate pre-processing (e.g., Canny edge detection or grid overlays), our Perception Agent directly feeds the raw engineering diagram into the VLM. This end-to-end strategy leverages the model's internal world knowledge to resolve visual ambiguities---such as distinguishing a ``dimension line'' from a ``structural beam''---that heuristic CV algorithms often misclassify.

\paragraph{Constraint-Aware Coordinate Estimation.} Extracting precise coordinates from unscaled raster images is ill-posed from a physical perspective. VFEAgent addresses this via a \textit{semantic estimation} strategy. When dimension annotations are present, the VLM infers the pixel-to-metric scale to establish the global coordinate system. In the absence of explicit dimensions, the system prioritizes \textit{topological consistency} (relative nodal positions) over absolute metric accuracy, ensuring that the generated finite element mesh remains mathematically valid for normalized analysis.

\paragraph{Hierarchical Validation Protocol.} To ensure that the generated IR is not just syntactically valid but also physically solvable, the Validation Agent enforces a four-level audit protocol derived from expert heuristics:
\begin{itemize}
    \item \textbf{L1 Schema Integrity:} Checks strictly against the JSON schema definitions (e.g., valid float types, non-empty arrays).
    \item \textbf{L2 Physical Stability:} Verifies static determinacy. For example, it detects ``rigid body motion'' risks by ensuring that constrained Degrees of Freedom (DoFs) cover both translation and rotation for frame structures.
    \item \textbf{L3 Engineering Plausibility:} Audits parameter ranges (e.g., rejecting negative Poisson's ratios or unrealistically high Young's modulus for standard steels).
    \item \textbf{L4 Visual Consistency:} Performs a reverse-check to ensure that all annotated dimensions in the input image are represented in the extracted topology.
\end{itemize}
Any violation at these levels triggers an immediate feedback loop to the Reasoning Agent before the IR is finalized.

\paragraph{Monotonic Completeness \& Context Pinning.} To circumvent context drift in multi-turn dialogues, we implement an \textit{Explicit Context Manager}. Critical engineering constraints (e.g., ``Topology Optimization Mode'') are pinned to the system prompt, preventing eviction during token window sliding. Furthermore, we enforce a \textit{Non-degenerative Acceptance Criterion}: updates to $Y$ are rejected if they result in information entropy loss (e.g., overwriting valid nodes with empty sets) unless explicitly justified by the Validation Agent.

\subsection{Stage-B: Verification-Driven Synthesis}
Stage-B addresses the ``Execution Gap'' inherent in LLM-generated code through a verification-driven synthesis protocol, as outlined in Algorithm~\ref{alg:text2abaqus}.

% --- ALGORITHM 1 (Added \small) ---
\begin{algorithm}[t]
\small
\caption{Reflexive Synthesis with Neural-Symbolic Handover}
\label{alg:text2abaqus}
\textbf{Input}: IR $Y$, Max Retries $K$, Memory $\mathcal{H}$ \\
\textbf{Output}: Valid Script $S^*$, ODB Result $R$
\begin{algorithmic}[1]
\STATE $Lessons \leftarrow \emptyset$
\STATE $S \leftarrow \text{LLM}_{\text{Gen}}(Y, \mathcal{H})$ \COMMENT{Conditioned on memory}
\FOR{$k = 0$ \textbf{to} $K$}
    \STATE $E_{static} \leftarrow \text{PreflightCheck}(S)$ \COMMENT{AST Analysis}
    \IF{$E_{static} \neq \emptyset$}
        \STATE $Lessons \leftarrow \text{Reflect}(E_{static})$
        \STATE $S \leftarrow \text{LLM}_{\text{Repair}}(S, Lessons)$
        \STATE \textbf{continue}
    \ENDIF
    \STATE \COMMENT{Execute in isolated sandbox}
    \STATE $R, E_{runtime} \leftarrow \text{ExecuteSubprocess}(S, \text{env})$
    \IF{$E_{runtime} = \emptyset$ \textbf{and} $\text{Verify}(R)$}
        \STATE $\text{UpdateMemory}(\mathcal{H}, S)$ \COMMENT{Index success skill}
        \STATE \textbf{return} $S, R$
    \ELSE
        \STATE $Lessons \leftarrow Lessons \cup \text{Debugger}(E_{runtime})$
        \STATE $S \leftarrow \text{LLM}_{\text{Repair}}(S, Lessons)$
    \ENDIF
\ENDFOR
\STATE $S_{det} \leftarrow \text{DeterministicFallback}(Y)$ \COMMENT{Symbolic Handover}
\STATE $R \leftarrow \text{ExecuteSubprocess}(S_{det}, \text{env})$
\STATE \textbf{return} $S_{det}, R$
\end{algorithmic}
\end{algorithm}

\paragraph{Preflight Static Verification.} 
Prior to execution, the generated script $S$ undergoes rigorous static analysis based on the Python Abstract Syntax Tree (AST). We enforce domain-specific constraints to intercept fatal errors, specifically: (1) \textit{Lifecycle Integrity}, which verifies the presence of essential \textbf{execution triggers} and \textbf{process termination signals} to prevent idle runs; and (2) \textit{API Safety}, which prohibits \textbf{unsafe kernel state manipulations} (e.g., deletion of protected root containers) that could compromise the solver environment.

\paragraph{Sandboxed Execution and Artifact Management.} 
Executing arbitrary LLM-generated code entails significant security and state-management risks. VFEAgent implements a strict \textit{Run Isolation} protocol, where each task is encapsulated within a dynamically allocated workspace. The generator is explicitly prompted to enforce \textbf{directory isolation} and redirect all \textbf{simulation artifacts} (including binary databases and status logs) to this secluded path. This mechanism prevents file collisions during parallel batch processing and ensures that output data is strictly associated with its parent task.

\paragraph{Reflexive Debugging with Experience Replay.} 
Upon runtime failure, the Debugger Agent synthesizes structured corrective feedback. We maintain a persistent \textit{Experience Replay Buffer} $\mathcal{H}$. For each instance, the system retrieves the top-$k$ isomorphic failure/success patterns from $\mathcal{H}$ based on error signature similarity. This retrieval mechanism enables the agent to preemptively resolve recurrent issues (e.g., version-specific syntax incompatibilities) by leveraging historical insights.

\paragraph{Schema-Gated Topology Optimization.} 
To accommodate advanced design synthesis beyond standard forward analysis, we incorporate a \textbf{Schema-Gated Branching} mechanism. Triggered by an explicit optimization intent (encoded via a dedicated semantic flag within the IR), the synthesis engine redirects the generation flow to a specialized protocol that instantiates the requisite \textbf{optimization task structures} and \textbf{solver process definitions}. To counteract the convergence instability typical of zero-shot generation, the system injects \textbf{Robust Initialization Priors} (e.g., imposing conservative volume constraints). These priors serve as operational baselines, guaranteeing solver stability while remaining adaptable to explicit user constraints extracted from the prompt.

\paragraph{Neural-Symbolic Handover (Deterministic Fallback).} 
Recognizing the stochastic nature of LLMs, we introduce a deterministic fallback mechanism. If the reflexive loop exhausts the retry budget $K$, the system triggers a \textit{Symbolic Handover}, reverting to a rule-based template engine that strictly maps the IR $Y$ to a guaranteed-executable script. This hybrid architecture ensures a baseline executability for canonical structural problems, effectively acting as a fail-safe.

% \paragraph{Implementation.}
% Our system utilizes a hybrid Python runtime (v3.9 for agents, v2.7 for Abaqus kernel). Detailed hardware specifications and structural signature retrieval algorithms are provided in \textbf{Appendix \ref{sec:appendix_details}}.
% \paragraph{Software Runtime.}
% Our system utilizes a hybrid Python runtime (v3.9 for agents, v2.7 for Abaqus kernel). Additional implementation details regarding the structural signature retrieval algorithm are provided in \textbf{Appendix \ref{sec:appendix_details}}.

% ==================================================================
% SECTION 4: EXPERIMENTS (Full Content with Fixed Floats)
% ==================================================================
\section{Experiments}
\subsection{Experimental Setup}
\paragraph{Implementation Details.} 
We implement the multi-agent system using a hybrid Python runtime (v3.9 for agents, v2.7 for Abaqus kernel). For comprehensive specifications regarding the \textbf{hardware environment}, \textbf{software dependencies}, and the \textbf{structural signature retrieval algorithm}, please refer to \textbf{Appendix \ref{sec:appendix_implementation}}.
\paragraph{Benchmarks and Baselines.}
We evaluate VFEAgent on a comprehensive suite of engineering problems, categorized into two modes: (1) \textit{Standard Forward Analysis}, encompassing trusses, frames, and plates under diverse boundary conditions; and (2) \textit{Topology Optimization}, focusing on compliance minimization subject to volume constraints (e.g., volfrac $  \le 0.5  $). We compare our approach against state-of-the-art multimodal LLMs, including GPT-4o, GPT-5 (Preview), Gemini-3-Pro, and Qwen-3-Max, each prompted with expert-crafted Chain-of-Thought (CoT) instructions.
\paragraph{Metrics.}
Performance is quantified using the following metrics: (1) \textit{Schema Validity}, measuring the adherence of the intermediate output to the strict FEM JSON schema; (2) \textit{Overall Perception Score}, a composite metric that averages node detection accuracy and connectivity F1-score; and (3) \textit{Execution Success Rate}, defined as the percentage of generated scripts that successfully complete the full Abaqus job lifecycle and yield a readable ODB file.

\subsection{Comparative Analysis}
\paragraph{Structural Interpretation (Stage-A).}
Table~\ref{tab:main_results} underscores the ``semantic gap'' inherent in foundation models. While general-purpose VLMs such as \textit{Gemini-3-Pro} demonstrate superior raw pixel-level perception (Node Acc.~0.815), they falter in mapping visual features to rigorous engineering schemas (Schema Validity 41.7\%). VFEAgent addresses this limitation through its multi-agent ReAct loop, which enforces logical consistency constraints (e.g., ensuring that every element connects exactly two existing nodes), thereby attaining a superior validity score of 90.0\%.

% --- TABLE 1: Fixed Format & Caption Bottom ---
\begin{table}[t]
\centering
\small
\setlength{\tabcolsep}{3.5pt}
\begin{tabular}{lccccc}
\toprule
\textbf{Model} & \textbf{Schema} & \textbf{Node} & \textbf{Conn.} & \textbf{BC} & \textbf{Overall} \\
 & \textbf{Valid.} & \textbf{Acc.} & \textbf{F1} & \textbf{Det.} & \\
\midrule
\textbf{VFEAgent} & \textbf{0.900} & \textbf{0.815}& \textbf{0.648}& \textbf{0.600} & \textbf{0.704} \\
Gemini-3-Pro & 0.417 & 0.775& 0.577& 0.458 & 0.639 \\
GPT-5 & 0.333 & 0.756 & 0.609 & 0.542 & 0.610 \\
GPT-4o & 0.583 & 0.750 & 0.488 & 0.500 & 0.596 \\
Gemini-3-Flash & 0.333 & 0.752 & 0.576 & 0.458 & 0.583 \\
Grok-4 & 0.545 & 0.582 & 0.500 & 0.500 & 0.503 \\
Qwen-3-Max & 0.091 & 0.127 & 0.024 & 0.136 & 0.128 \\
\bottomrule
\end{tabular}
\caption{Interpretation performance (Stage-A). VFEAgent achieves the highest Schema Validity (90.0\%), significantly outperforming baselines.}
\label{tab:main_results}
\end{table}

\paragraph{Execution Robustness (Stage-B).}
The ``execution gap'' is further examined in Table~\ref{tab:script_quality}. A key insight is that superior linguistic proficiency does not necessarily equate to engineering accuracy. For example, \textit{GPT-5} achieves a score of 1.0 on Lifecycle (accurately handling job submission) but 0.0 on Preflight Safety, largely attributable to the generation of unsafe API calls. VFEAgent's neuro-symbolic handover protocol guarantees 100\% success by reverting to deterministic templates when the probabilistic generator fails to converge.
% --- Critical Clarification Paragraph (Defense against ``Cheating'') ---
It is essential to elucidate the role of the deterministic fallback mechanism in our evaluation. Although conceived as a vital safeguard for real-world industrial applications, \textbf{this mechanism was inactive (0\% activation rate) across all experimental benchmarks presented in Table~\ref{tab:script_quality}.} The 100\% execution success rate was attained solely through the neural agents employing the reflexive debugging loop ($  K \le 3  $), thereby affirming the inherent robustness of our neuro-symbolic architecture without dependence on rule-based templates in these standard scenarios.

% --- TABLE 2: Fixed Format & Caption Bottom ---
\begin{table}[t]
\centering
\small
\setlength{\tabcolsep}{2pt}
\begin{tabular}{l|c|cc|ccc}
\toprule
 & \textbf{Exec.} & \multicolumn{2}{c|}{\textbf{Safety}} & \multicolumn{3}{c}{\textbf{Physical Comp.}} \\
\textbf{Model} & \textbf{Preflight} & \textbf{Life.} & \textbf{Help.} & \textbf{BC} & \textbf{Load} & \textbf{Avg.} \\
\midrule
Qwen-3-Max & 0.50 & \textbf{1.0} & \textbf{1.0} & 0.69 & 0.37 & 0.53 \\
Grok-4 & 0.50 & \textbf{1.0} & \textbf{1.0} & 0.69 & 0.37 & 0.53 \\
Gemini-3-Pro & 0.25 & 0.5 & \textbf{1.0} & 0.38 & 0.30 & 0.34 \\
GPT-5 & 0.00 & \textbf{1.0} & 0.75 & \textbf{0.81} & \textbf{0.39} & \textbf{0.60} \\
\midrule
\textbf{Ours} & \textbf{1.00} & \textbf{1.0} & \textbf{1.0} & \textbf{1.00} & \textbf{1.00} & \textbf{1.00} \\
\bottomrule
\end{tabular}
\caption{\textbf{Granular Analysis (Stage-B).} Breakdown of generation quality. \textbf{Ours} guarantees 100\% compliance across all safety and physical metrics.}
\label{tab:script_quality}
\end{table}

% --- FIGURE 3: Case Study ---
\begin{figure*}[t]
    \centering
    % 请确保使用双栏宽度的图片文件
    \includegraphics[width=0.95\textwidth]{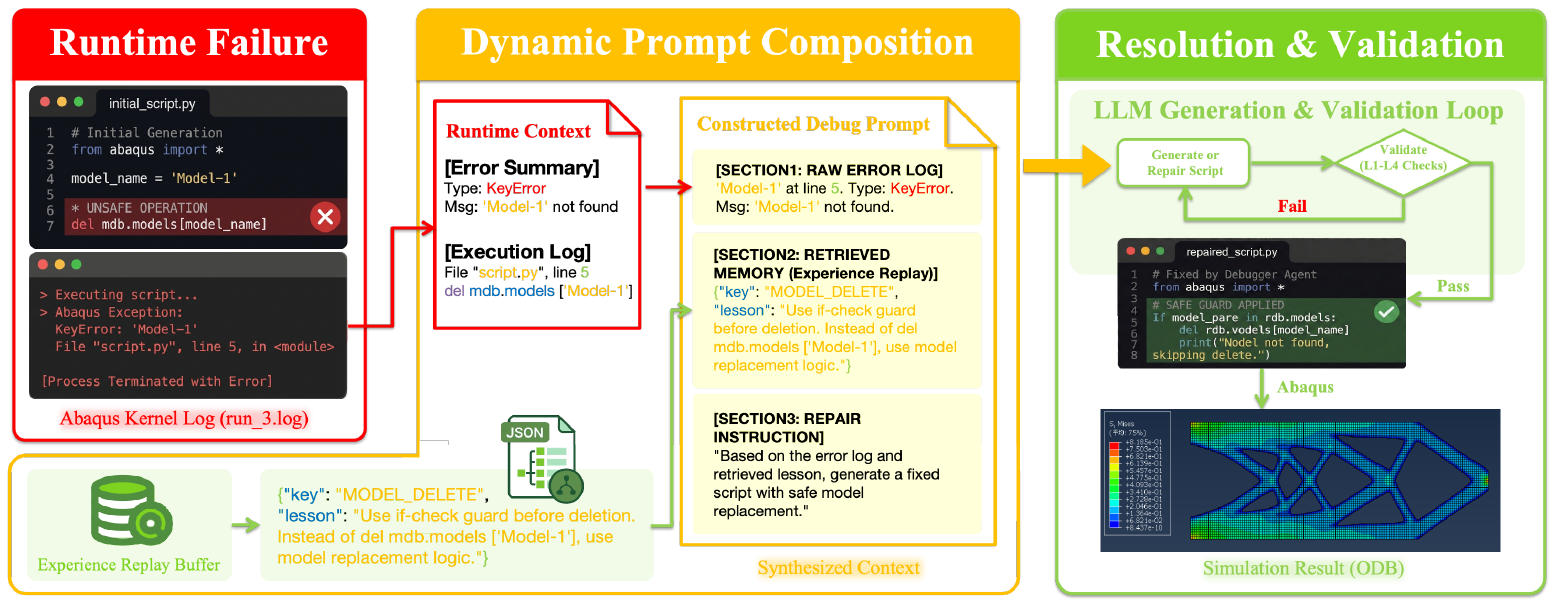} 
    \caption{\textbf{Case Study: Autonomous Recovery via Composite Context Debugging.} 
    \textbf{(A) The Crash:} The script attempted an unsafe root model deletion, triggering a \texttt{KeyError}.
    \textbf{(B) Context Assembly (The "Hard" Part):} Instead of simple error matching, the Debugger Agent constructs a composite prompt. It aggregates the \textbf{Structured Error Summary} (for high-level intent) and the \textbf{Raw Execution Log} (for line-level localization). Simultaneously, it retrieves a specific \textit{Experience Lesson} (JSON) to ground the repair in domain rules.
    \textbf{(C) Verified Fix:} Armed with both the specific traceback and the domain constraint, the Repair Agent synthesizes a precise \texttt{try-except} guard, successfully resolving the crash.}
    \label{fig:reflexive_case}
\end{figure*}

\subsection{Taxonomy of LLM Engineering Failures}
Beyond success rates, we categorize the dominant failure modes of foundation models in CAE tasks, derived from the error logs of Experiment 2.
\paragraph{Type I: Lifecycle Blindness (45\% of failures).}
Foundation models often treat the script as a static description rather than a dynamic process. A frequent error involves defining the entire model correctly but omitting the \texttt{job.submit()} or \texttt{waitForCompletion()} calls. This results in an ``idle run'' that terminates without triggering the solver kernel, thereby producing no ODB data.
\paragraph{Type II: Contextual API Hallucination (30\% of failures).}
Models often invent non-existent methods by conflating different object levels. For example, we observed models calling \texttt{mesh.EdgeArray(...)}---a method that does not exist in the Abaqus API---instead of accessing edges via the Part object. This reflects a lack of fine-grained knowledge about the hierarchical Abaqus Object Model (AOM).
\paragraph{Type III: Unsafe State Manipulation (25\% of failures).}
To ensure a ``clean slate,'' models frequently attempt to execute \texttt{del mdb.models['Model-1']}. In the Abaqus CAE environment, `Model-1' is often the protected root container, and deleting it causes an immediate kernel crash. This nuance is rarely documented in public training corpora, rendering it a common zero-shot failure mode.
\subsection{Qualitative Analysis: Self-Correction in Action}
Figure~\ref{fig:reflexive_case} presents a detailed execution log, demonstrating the system's robust debugging capability.
\paragraph{The Runtime Failure (Panel A).}
The Executor Agent attempted to sanitize the environment using \texttt{del mdb.models['Model-1']}. While valid in standard Python, this operation violates the Abaqus kernel's protection of the root node, resulting in a fatal \texttt{KeyError}.
\paragraph{Dynamic Context Composition (Panel B).}
The core of our debugging engine lies not merely in retrieving a fix, but in constructing an \textbf{information-complete context} for the LLM. As detailed in the prompt protocols (Appendix B), the Debugger Agent performs a multi-source aggregation:
\begin{enumerate}
\item \textbf{Log Parsing:} It captures the full \textit{execution traceback} to localize the fault line (Line 45) and generates a structured \textit{error summary} (e.g., Invalid Dictionary Key Access'').     \item \textbf{Knowledge Retrieval:} It uses the error signature to query the Experience Replay Buffer, retrieving the crucial domain constraint: \textit{Root model deletion is forbidden.''}
\end{enumerate}
This composite input ensures that the model understands not just \textit{what} failed (the error), but also \textit{where} (the log) and \textit{how to fix it} (the lesson).
\paragraph{Successful Resolution (Panel C).}
Guided by this rich context, the Repair Agent implemented a defensive \texttt{try-except} block. The script successfully converged (15 iterations), confirming that feeding raw runtime logs alongside retrieved knowledge is essential for resolving ``Type III'' engineering failures.
\subsection{Discussion and Limitations}
While VFEAgent demonstrates strong capabilities in automating standard structural analysis, we identify two primary limitations:
\paragraph{Dependence on Visual Fidelity.}
The system's performance is currently bounded by the resolution of the upstream VLM. In cases of low-resolution blueprints where dimension lines overlap with structural members, the Perception Agent may hallucinate connectivity. Future work will integrate vector-graphics parsing (SVG) to bypass raster-level ambiguity.
\paragraph{3D Reconstruction Ambiguity.}
Extracting full 3D depth from a single 2D orthographic view remains an ill-posed problem. Our current ``Constraint-Aware'' estimation relies on standard engineering conventions (e.g., extruding 2D profiles). Handling complex non-manifold 3D geometries would require multi-view reasoning capabilities, which represents a direction for our next-generation framework.

% ==================================================================
% SECTION 5: CONCLUSION
% ==================================================================
\section{Conclusion}

In this paper, we introduce VFEAgent, a neuro-symbolic multi-agent framework that automates the end-to-end workflow of finite element analysis. By decomposing the complex modeling task into a ReAct-driven visual perception stage and a verification-first synthesis stage, VFEAgent effectively bridges the ``semantic gap'' in interpreting engineering diagrams and the ``execution gap'' in generating rigorous simulation scripts.

Our empirical evaluations indicate that large language models, when augmented with hierarchical validation and reflexive debugging, can achieve near-industrial-grade reliability. VFEAgent demonstrated robust performance across the evaluated structural benchmarks, yielding valid simulation outcomes solely through neural self-correction ($K \le 3$) in the tested scenarios. Moreover, the automated workflow exhibited a substantial efficiency gain, reducing modeling latency from typical manual durations (minutes) to near-real-time execution (seconds), thereby underscoring its potential to streamline iterative design processes in computational engineering.

Future work will address the inherent limitations of 2D raster inputs by incorporating 3D-native perception modalities (e.g., point clouds or STEP files) to manage complex non-manifold geometries. Additionally, we plan to extend the framework's solver-agnostic intermediate representation to support open-source engines such as OpenSees and CalculiX, thereby further democratizing access to autonomous simulation technologies.

\bibliographystyle{named}
\bibliography{ijcai26}

\appendix
\newpage
\section{Implementation Details}
\label{sec:appendix_implementation}

\paragraph{Hardware \& Computing Environment.} 
To ensure a rigorous evaluation of inference latency, all baselines and our VFEAgent system were evaluated on the same high-performance workstation. The testbed is equipped with dual \textbf{AMD EPYC 9654} processors (384 threads) and 512 GB of RAM. The simulation engine is powered by \textbf{Simulia Abaqus 2022} running on a Linux environment. This standardized hardware setup eliminates performance variances caused by differing compute resources.

\paragraph{Software Architecture.} 
The multi-agent system is orchestrated via the LangChain framework. To bridge the compatibility gap between modern AI agents and legacy CAE software, we employ a dual-environment architecture: the agents operate in a Python 3.9 runtime, while the simulation execution and ODB extraction are handled by an isolated Python 2.7 sandbox strictly compatible with the Abaqus Scripting Interface.

\paragraph{Structural Signature Retrieval.} 
Instead of relying on opaque vector embeddings, we implement a transparent \textbf{Heuristic Structural Signature} matching algorithm for experience replay. We define a model's signature as a tuple $S = (N_{node}, N_{elem}, \{T_{sec}\}, N_{load}, N_{bc})$. During runtime debugging, the system retrieves historical success cases where the node-to-element density ratio falls within a calibrated tolerance interval $[0.5, 2.0]$. This engineering-centric metric ensures that retrieved lessons are topologically relevant to the current failure mode.

\end{document}